# Investigating self-supervised, weakly supervised and fully supervised training approaches for multi-domain automatic speech recognition: a study on Bangladeshi Bangla

Ahnaf Mozib Samin[1,2], M. Humayan Kobir[1], Md. Mushtaq Shahriyar Rafee[1], M. Firoz Ahmed[1], Mehedi Hasan[1], Partha Ghosh[1], Shafkat Kibria[1,3], and M. Shahidur Rahman[1]

[1]Department of Computer Science and Engineering, Shahjalal University of Science and Technology, Bangladesh

[2]Faculty of Information & Communication Technology, University of Malta, Malta

[3]Department of Computer Science and Engineering, Leading University, Bangladesh

Despite huge improvements in automatic speech recognition (ASR) employing neural networks, ASR systems still suffer from a lack of robustness and generalizability issues due to domain shifting. This is mainly because principal corpus design criteria are often not identified and examined adequately while compiling ASR datasets. In this study, we investigate the robustness of the state-of-the-art transfer learning approaches such as self-supervised wav2vec 2.0 and weakly supervised Whisper as well as fully supervised convolutional neural networks (CNNs) for multi-domain ASR. We also demonstrate the significance of domain selection while building a corpus by assessing these models on a novel multi-domain Bangladeshi Bangla ASR evaluation benchmark — BanSpeech, which contains approximately 6.52 hours of human-annotated speech and 8085 utterances from 13 distinct domains. SUBAK.KO, a mostly read speech corpus for the morphologically rich language Bangla, has been used to train the ASR systems. Experimental evaluation reveals that self-supervised cross-lingual pre-training is the best strategy compared to weak supervision and full supervision to tackle the multi-domain ASR task. Moreover, the ASR models trained on SUBAK.KO face difficulty recognizing speech from domains with mostly spontaneous speech. The BanSpeech will be publicly available to meet the need for a challenging evaluation benchmark for Bangla ASR.

## 1 INTRODUCTION

Transfer learning has become the state-of-the-art approach in natural language and speech processing [Devlin et al. 2018; Baevski et al. 2020]. By pre-training a model with a huge amount of unlabelled speech data in a self-supervised fashion and then fine-tuning it with a much smaller dataset, the performance of the ASR models can be significantly improved. Despite the improvements brought by self-supervised training and wav2vec 2.0 [Baevski et al. 2020], some studies indicate that speech models pre-trained in a supervised way for numerous domains are more capable of ensuring robust performance [Narayanan et al. 2018; Likhomanenko et al. 2020; Chan et al. 2021]. However, collecting thousands of hours of high-quality supervised speech data is a cumbersome and expensive task. To this end, scaling weakly supervised pre-training is investigated for speech processing and found to be the state-of-the-art for English ASR even in zero-shot transfer learning without additional fine-tuning [Radford et al. 2022]. While creating a weakly supervised speech dataset, the human-validated annotation is not performed and data is curated using an automated pipeline. However, the benefit of weak supervision is that it is comparatively easier to scale the dataset and as for Whisper, approximately 680K hours of multilingual data is curated in this way [Radford et al. 2022]. Yet, the zero-shot performance of Whisper for low-resource languages is not up-to-the-mark. The authors hypothesize that fine-tuning might improve the results which is yet to be investigated for many languages. It still remains an open question how robust the current fully supervised, self-supervised, and weakly supervised speech models are for multiple-domain speech recognition.

Bangla is a morphologically-rich language from the Indo-Aryan language sub-group. Kibria et al. have developed SUBAK.KO, an annotated speech corpus for speech recognition research comprising 241 hours of Bangladeshi Bangla speech data, to address the dearth of annotated speech datasets in Bangla [Kibria et al. 2022]. SUBAK.KO contains 229 hours of clean read speech and 12 hours of broadcast speech utterances. The recording scripts are collected from 40 text domains, including conversations, sports, news, poetry, letters, etc., following the reception and production criteria for the text domains to build the corpus [Atkins et al. 1992]. According to a cross-dataset evaluation, SUBAK.KO is a more balanced corpus with respect to regional accents and other types of speaker variability compared to another large-scale Bangladeshi Bangla speech corpus, LB-ASRTD, when evaluated on clean read speech test sets [Kibria et al. 2022; Kjartansson et al. 2018]. The read speech test sets from standard datasets often cannot evaluate the robustness of ASR applications since the same domains are also included in the training set [Geirhos et al. 2020]. Thus, it is yet uncertain how well an ASR model based on SUBAK.KO performs in a range of domains, particularly those that include the majority of spontaneous speech. Human verbal communication consists primarily of spontaneous speech with some background noise, and past research indicates significant acoustical and linguistic variations between read speech and spontaneous speech [Nakamura et al. 2008].

Motivated by the above-mentioned issues with regard to multi-domain ASR, this paper focuses on the following research questions: 1) In terms of robustness and generalizability, how do the state-of-the-art fully supervised, self-supervised, and weakly supervised speech recognition models perform for multiple domains, especially those that contain mostly spontaneous speech? 2) Do the 40 text domains considered while developing SUBAK.KO meet all of the reception and production criteria for developing an ASR corpus that can ultimately lead to improved performance in challenging real-world conditions, such as noisy, spontaneous, and multi-talker environments?

The contributions of this paper are as follows:

- We present BanSpeech, a novel multi-domain Bangladeshi Bangla ASR evaluation benchmark consisting of 6.52 hours of human-validated speech data and 8085 utterances from 13 domains collected from YouTube and manually transcribed by human annotators. This speech dataset consists primarily of spontaneous speech from all domains, with the exception of *audiobooks* and *biography* domains, which comprise read speech. Moreover, we collect 80 minutes of additional data containing dialectal speeches from the seven major divisions of Bangladesh, making BanSpeech 7.7 hours long. The latter part, however, has not been validated under human-supervision.
- We explore a fully supervised deep CNN, a self-supervised wav2vec 2.0 XLS-R [Babu et al. 2021], and a weakly supervised Whisper model [Radford et al. 2022] to investigate their robustness on multiple domains using BanSpeech evaluation set. We train the CNN model from scratch using SUBAK.KO. For the wav2vec 2.0 and Whisper, we choose the cross-lingual pre-trained models and the same SUBAK.KO dataset for fine-tuning. All of the models are evaluated on the multi-domain BanSpeech and SUBAK.KO test set.
- Through a comprehensive evaluation of ASR models trained/fine-tuned on SUBAK.KO, we shed light on the significance of domain selection for the development of a speech dataset. In this regard, both read and spontaneous speeches from BanSpeech are used to evaluate them.
- We report our findings on deep CNN by experimenting with the number of convolutional layers, applying several normalization techniques, trying out some input feature extraction methods from audio signals, and observing the effect of the number of Mel Frequency Cepstral Coefficients (MFCCs).

The rest of the paper is structured as follows: background is provided in section 2. In section 3, the preparation of the BanSpeech is described. We discuss the experimental setup in section 4. Results are shown and discussed in section 5. We present the limitations of our work in section 6. Section 7 concludes the article and provides the future direction.

## 2 BACKGROUND

**Related work in Bangla** While prior to 2018, Bangla ASR research was limited to only isolated words and digit recognition using small datasets, some work has been done on Bangla large vocabulary continuous speech recognition (LVCSR) later on [Sultana et al. 2021]. Amin et al. examined Deep Neural Network-Hidden Markov Model (DNN-HMM) and Gaussian Mixture Model-Hidden Markov Model (GMM-HMM) based techniques on a relatively small and speaker-independent Shruti corpus (21.64 hours) [Al Amin et al. 2019; Das et al. 2011]. According to their findings, a larger data set is necessary for the DNN-HMM method to outperform the GMM-HMM method. Sumit et al. implemented the Recurrent Neural Network-based Deep Speech 2 architecture on the non-public 300-hour-long "Socian" Bangla telephone conversation-based dataset [Sumit et al. 2018]. In addition to Socian, they also added 50 hours from the Bangla Babel corpus. Developed in 2016, Bangla Babel is likewise a telephone conversation-based speech corpus, comprising 215 hours of speech [IARPA 2020]. However, Babel has West Bengal accented speech that is distinct from the Bangladeshi Bangla accent [Kibria et al. 2022]. Ahmed et al. prepared 960 hours of broadcast Bangla speech corpus by transcribing speech data in an automated way with pre-trained ASR models [Ahmed et al. 2020]. The corpus is not publicly accessible, and the authors focused on developing an algorithm to iteratively construct speech corpora in their work. However, the objective of our work is to empirically evaluate the ASR models trained with three distinct approaches and the SUBAK.KO dataset on diverse domains to find out whether performance can vary across different domains.

Samin et al. evaluated the quality of a large-scale publicly available LB-ASRTD corpus (229 hours) using deep learning-based approaches by conducting character-wise error analysis [Samin et al. 2021]. They also found a deep CNN-based acoustic model and a 5-gram Markov Language Model (LM) to be capable of achieving a lower word error rate (WER) on LB-ASRTD. In this study, we also use a deep CNN-based model while utilizing a higher number of MFCCs during the input feature extraction and introducing layer normalization in each convolution layer. Based on an acoustic study on a regional accented speech and the character-wise error analysis on LB-ASRTD, the requirement of a new corpus with more speaker variability and character-wise well-balancedness was recommended [Kibria et al. 2020; Samin et al. 2021]. Therefore, Kibria et al. developed the 241-hour long publicly available Bangladeshi Bangla SUBAK.KO corpus with the aim of addressing the above-mentioned issues of LB-ASRTD [Kibria et al. 2022]. The Bengali Common Voice Speech dataset with over 400 hours of crowd-sourced data has been made available on the Mozilla Common Voice Platform, and the campaign to address the scarcity of Bangla speech datasets is ongoing [Alam et al. 2022]. Although there are now some large annotated ASR corpora for Bangla, there is no comprehensive ASR evaluation benchmark in this language that can categorically investigate a model on a variety of domains, dialects, and speech types from multiple speakers.

Shahgir et al. fine-tuned a wav2vec 2.0 model on the Bengali Common Voice Speech dataset and presented a promising performance for Bangla ASR [Shahgir et al. 2022]. They choose the wav2vec 2.0 model pre-trained on 53k hours of cross-lingual speech data in a self-supervised way [Conneau et al. 2020]. However, there also exists wav2vec 2.0 XLS-R model pre-trained on a much larger amount of data (436K hours) and Bengali is among the languages which were considered during pre-training [Babu et al. 2021]. To the best of our knowledge, no previous work has investigated the Whisper [Radford et al. 2022] for Bangla ASR.

**wav2vec 2.0** Wav2vec 2.0 is pre-trained in a self-supervised way by masking input frames in the latent space and using a contrastive loss function, the model learns the inherent representations from speech [Baevski et al. 2020]. These representations have now been used in numerous speech processing tasks including ASR, language identification, keyword spotting, speaker verification, speech translation, etc [Fan et al. 2020; San et al. 2021; Li et al. 2020]. wav2vec 2.0 takes raw audio as input and passes it to a feature encoder consisting of convolutional blocks. A quantization module is used to transform the output of the feature encoder from the continuous space into discrete space and a context network with Transformer blocks is trained with a contrastive objective.

Following the success of monolingual wav2vec 2.0, a larger cross-lingual wav2vec 2.0 model named XLS-R is released [Babu et al. 2021]. XLS-R has been pre-trained on 436K hours of data from 128 languages. There are three versions of this model with 300 million, 1 billion and 2 billion parameters. During the pre-training stage, several Indo-Aryan languages have been included in XLS-R. More precisely, there are 100 hours of Bengali training data from the common voice (CV) dataset which is used for pre-training the XLS-R. It has been shown that when a model is pre-trained on some closely-related languages, then it substantially improves the ASR performance after fine-tuning [Babu et al. 2021]. For this reason, XLS-R is found to be extremely useful to build ASR applications for many low-resource and mid-resource languages.

**Whisper** Contrary to the state-of-the-art method of self-supervised learning, Whisper is pre-trained on 680K hours of speech data adopting a weakly supervised pre-training method [Radford et al. 2022]. Instead of ensuring gold-standard human-validated transcripts, they collected audio data and corresponding transcripts from the internet. Since these transcripts are often noisy (e.g. ASR generated transcripts, etc), they utilized several automated filtering strategies. The training datasets for Whisper are prepared without human supervision. Whisper is pre-trained with a standard encoder-decoder Transformer [Vaswani et al. 2017]. The detailed architecture is described in their paper [Radford et al. 2022]. Similar to wav2vec 2.0 XLS-R [Babu et al. 2021], Whisper is also pre-trained cross-lingually in 97 languages and contains several versions including tiny, base, small, medium and large with 39 million, 74 million, 244 million, 769 million, 1550 million of parameters, respectively.

While Whisper sets a state-of-the-art on English LibriSpeech benchmark, the zero-shot performance of other languages using Whisper is substantially low. The reason lies in the fact that Whisper is pre-trained mostly on English data and the other 96 languages have less than 1000 hours of data. In the case of Bengali, there are only 1.3 hours of annotated speech data used for pre-training and the WER is more than 80% in a zero-shot setting. However, the authors hypothesize that the Whisper performance for low-resource languages can be improved if fine-tuned on a specific language with more data instead of performing zero-shot transfer.

**Multi-domain ASR** Hsu et al. conducted extensive experiments on domain shift in self-supervised pre-training [Hsu et al. 2021]. They found that adding diverse domains in pre-training data helps improve the robustness of the model and the model can recognize data not seen during training. Speech enhancement, data augmentation, and autoencoders are investigated for domain adaptation in distant speech recognition [Tang et al. 2018]. Domain adaptation for robust ASR is also studied adopting self-training [Khurana et al. 2021] and multi-task learning [Sun et al. 2017]. Kawakami et al. pre-trained 8000 hours of noisy multilingual data from diverse domains in an unsupervised way and showed that the learned representations are more robust to domain shifts [Kawakami et al. 2020]. To the best of our knowledge, no work has been done to investigate the robustness of self-supervised, weakly supervised and full supervised models by dissecting speech data into multiple domains.

**Table 1** Key statistics of BanSpeech

| Domain Type | Domain | Length | Number of samples | Vocabulary size |
|---|---|---|---|---|
| General domains | Television news | 30.3 min | 571 | 1786 |
| | Parliament speech | 30.0 min | 585 | 1585 |
| | Audio books | 30.3 min | 955 | 2339 |
| | Drama series | 30.5 min | 514 | 1628 |
| | Class lecture | 30.2 min | 397 | 1288 |
| | Political talk show | 30.0 min | 813 | 1798 |
| | Celebrity Interview | 30.5 min | 561 | 1752 |
| | Documentary | 30.2 min | 615 | 1773 |
| | Kids' voice | 30.3 min | 321 | 1497 |
| | Medicine | 31.2 min | 704 | 1294 |
| | Biography | 28.2 min | 657 | 1739 |
| | Kids' Cartoon | 29.6 min | 660 | 1439 |
| | Sports | 30.3 min | 732 | 1784 |
| Dialectal domains | Barisal | 10.8 min | 129 | 688 |
| | Chittagong | 10.5 min | 154 | 589 |
| | Dhaka | 10.5 min | 189 | 690 |
| | Mymensingh | 8.6 min | 141 | 810 |
| | Noakhali | 10.2 min | 93 | 725 |
| | Rajshahi | 10.1 min | 229 | 543 |
| | Sylhet | 10.0 min | 199 | 503 |
| - | Total validated | 391.5 min or 6.52 hours | 8085 | 11630 |
| - | Total (including dialects) | 462.2 min or 7.7 hours | 9219 | 12807 |

# 3 PREPARATION of BanSpeech DATASET

*We collect speech data from the open-source platform YouTube. We consider 20 domains, namely television news, parliament speech, audio books, drama, class lecture, political talk show, interview, documentary, kids' voice, medical talk show, biography, sports, cartoon and 7 regional dialects from all the major divisions in Bangladesh such as Barisal, Chittagong, Dhaka, Mymensingh, Noakhali, Rajshahi,* and *Sylhet*. All the domains are annotated except for the 7 dialectal domains. The regional dialects are mainly obtained from region-specific dramas. There are essentially two types of talk shows we consider in this work: *political* and *medical* related. On political talk shows, politicians argue about political matters, while on medical talk shows, medical practitioners discuss medical-related topics and frequently employ scientific jargon. Some domains, such as *television news*, *parliamentary speeches*, and *documentaries*, incorporate both read and spontaneous speech, while others, such as *drama*, *class lectures*, *political talk shows*, *interviews*, *kid's voices*, and *medical talk shows*, consist predominantly of spontaneous speech. The *audio book* and *biographical* domains only contain read speech. For each of the domains, there are approximately 30 minutes of speech, except for the dialect domain, which has around 10 minutes of speech per dialect.

We download broadcast speech as waveform audio file format (WAV). We remove commas and brackets from the original audio file names and replace spaces with underscores to avoid potential errors. To make the corpus consistent, each audio file is then converted to a bitrate of 256 kilobits per second (kbps) and a 16 kilo Hertz (kHz) mono channel WAV file. After that, each audio file is automatically divided into smaller chunks based on silence intervals because they are initially too long to pass them to a neural network. We use a silence threshold of -40 decibels relative to full scale (dBFS) (consider it silent if quieter than -40 dBFS) and 0.2 seconds as the minimum length of silence (the shortest period of silence before a split can happen). Our dataset contains audio files of varying lengths. The lengths of the audio files range from 0.7 to 35 seconds.
The annotation process is conducted in three stages. After preparing the audio files, we first use the Google Speech-to-Text (STT) system to get the transcripts of those speeches [Google Bangla STT 2022]. Then, two professional native Bangladeshi human annotators with linguistic expertise manually correct the transcriptions. The transcripts are again corrected and finally verified by a new annotator. The corresponding text transcriptions are stored in plain text file format (.txt).

The verified annotated dataset contains 8085 utterances and 6.52 hours of speech with transcriptions. The dialectal domain is not annotated under human supervision since those utterances are extremely deviant from the standard Bangla, possess multiple issues related to annotation and require native annotators from the corresponding regions. Although the dialectal domains are not suitable to be included in an ASR evaluation set, they can be used for language/dialect identification tasks. The detailed statistics of BanSpeech is shown in Table 1.

# 4 EXPERIMENT SETUP

**Dataset** We train and evaluate our ASR models on SUBAK.KO which contains 241 hours of transcribed Bangladeshi Bangla speech data [Kibria et al. 2022]. This corpus has 229 hours of clean read speech and only 12 hours of broadcast speech. Clean read speeches are recorded from 33 native Bangladeshi Bangla male speakers, 28 native female speakers, and 2 second language (L2) speakers. The detailed description of this corpus can be found in the work of Kibria et al. [Kibria et al. 2022]. The same train, dev and test splits used in the original paper have been used in our study. The train, dev and test set contain 200.28 hours, 20.54 hours and 20.30 hours of speech data, respectively. SUBAK.KO train set is used to train our baseline CNN model from scratch and to fine-tune both wav2vec 2.0 and Whisper. For evaluation, we use the SUBAK.KO test set and BanSpeech.

**Baseline** We use a deep CNN model as our baseline in this study. For the CNN, three types of features are used as input to the acoustic model in different experiments: mel-frequency cepstral coefficients (MFCC), mel-frequency spectral coefficients (MFSC) and power-spectrum. MFCCs are a widely used input feature in ASR. MFSC refers to the log-energy directly computed from the mel-frequency spectral coefficients without applying Discrete Cosine Transform (DCT). Power-spectrum features have been used in acoustic modeling in speech recognition [Amodei et al. 2016]. Although 13 MFCCs are generally considered sufficient for extracting acoustic features, we experiment with different numbers of MFCCs to see its impact on WER using SUBAK.KO. Non-stationary speech signal is split into smaller windows or frames where it can be assumed as stationary. Different frame sizes and frame strides can affect the WER which we also investigate in this work.

There are several variations in the architecture for conducting numerous experiments. In the case of our baseline CNN model, MFCC feature vectors are fed into the CNN. Our baseline architecture consists of mainly convolutional layers and fully connected linear layers. Our architecture includes 20 convolutional layers each with a kernel size of 8*1. In the first convolutional layer, 18 MFCCs are mapped to an embedding space of size 256 and stride size 2 is used only for this layer. In the rest of the convolutional layers, there are 256 input channels and 256 output channels (feature maps) and feature extraction through convolution operation is done by 256 filters with stride size 1. Each of the convolutional layers is followed by ReLU activation function to add non-linearity to the network and dropout to address the over-fitting problem. The dropout probability is 0.1. Lastly, two fully-connected linear layers are attached for the classification of character-level tokens.

Apart from the baseline architecture, we train several acoustic models incorporating the batch norm [Ioffe and Szegedy 2015], layer norm [Ba et al. 2016] and weight norm techniques [Salimans and Kingma 2016]. After running numerous experiments, we obtain our best-performing CNN. We only use our best-performing CNN model to compare its performance with wav2vec 2.0 and Whisper on the BanSpeech. For training the acoustic model, we utilize the wav2letter++ speech processing toolkit [Pratap et al. 2018].

**Transfer learning** We fine-tune the wav2vec 2.0 XLS-R cross-lingual pre-trained model that has 300 million of trainable parameters [Babu et al. 2021]. As for the Whisper model, we choose the medium sized model with 769 million parameters. We use SUBAK.KO to fine-tune both of them. There exists standard HuggingFace implementations for wav2vec 2.0 and Whisper, which we make use of to run our experiments [Wolf et al. 2019].

**Greedy Decoding** Instead of beam search decoding with language model rescoring, we only perform greedy decoding using connectionist temporal classification (CTC). While rescoring with the help of a language model achieves better performance [Samin et al. 2021], we would like to observe the WERs and CERs of the acoustic models without any interference from the language model. The goal of this study is to investigate the robustness of several ASR systems across multiple domains and evaluate SUBAK.KO on both read and spontaneous speech. For this, we do not require language model rescoring.

All the experiments are conducted using a single GPU with 24 gigabytes (GB) of VRAM and a central processing unit (CPU) with 24 cores. Some of the major hyper-parameters of our CNN, wav2vec 2.0 and Whisper-based implementations are provided in Appendix A.

## 5 RESULTS & DISCUSSION

### 5.1 Evaluating robustness using BanSpeech

We present the results of deep CNN, wav2vec 2.0 and Whisper models, either trained from scratch (CNN) or fine-tuned (wav2vec 2.0 and Whisper) on SUBAK.KO in Table 2. We use our best-performing CNN with 20 convolutional layers, layer normalization in each convolutional block, and 21 MFCCs extracted from each frame. We show the performance across 13 domains from BanSpeech as well as using the SUBAK.KO test set. To evaluate the models, we use the most standard ASR metrics such as word error rate and character error rate. Moreover, we calculate the out-of-vocabulary (OOV) rates for each domain in reference to SUBAK.KO train and val sets. OOV rate of a domain is calculated from dividing the vocabulary size of that domain by the total vocabulary of the SUBAK.KO train and dev sets. For example, the *audiobooks* domain has a vocabulary of 2339 and the vocabulary of the SUBAK.KO train and dev sets is 37301. There are 406 words in that domain which are unavailable in SUBAK.KO. Dividing 406 (number of OOV words) by 2339 (vocabulary size) provides us with a 17.36% OOV rate. For each of the domains, the OOV rate is greater than 10% which implies the difficulty of building a robust ASR model on the morphologically rich language Bangla. Medicine domain has the highest OOV rate because of the academic jargon used in this domain. BanSpeech has a much higher OOV rate (30.83%) than SUBAK.KO test set (5.24%), which indicates that domain shifting introduces new vocabularies that are unseen during training.

For all the 13 domains, wav2vec 2.0 obtains substantially lower WERs compared to Whisper and our baseline CNN. Whisper performs better compared to CNN on each domain except for *biography*. Moreover, the difference of WERs between CNN and Whisper for the *audiobook* domain is marginal. It is worthwhile to mention that both *biography* and *audiobook* contain read speeches. From these findings, we can conclude that a self-supervised wav2vec 2.0 can recognize speeches from both read speech and spontaneous speech domains better than CNN and Whisper. A CNN, trained from scratch, is comparatively better with read speech samples, however, performs poorly on spontaneous speeches. The fact that CNN is trained solely with a read speech SUBAK.KO corpus might be the reason behind the poor performance on spontaneous speeches. This emphasizes the requirement for a spontaneous speech corpus for Bangla ASR. In the case of the Whisper that is pre-trained under weak supervision, the performance is not up-to-the-mark on Bangla even after fine-tuning with 200 hours of high-quality supervised data.

On the BanSpeech benchmark, wav2vec 2.0 achieves 40.62% WER while CNN and Whisper obtain 67.65% and 55.92% WERs, respectively. wav2vec 2.0 also sets a benchmark on the SUBAK.KO test set getting a WER of 7.56%. CNN and Whisper get 13.94% and 22.15% WERs on the SUBAK.KO test set. We argue that since BanSpeech contains a large amount of spontaneous speech, it makes it a challenging evaluation set for ASR models. These results also prove the fact that when ASR models are evaluated on the test set split from the same corpus, the performance cannot be properly evaluated as the same domains/speakers are seen during training.

**Table 2** CERs/WERs calculated for CNN, wav2vec 2.0 XLS-R and Whisper trained on SUBAK.KO. 13 distinct domains from BanSpeech and SUBAK.KO test set is used for evaluation. Out-of-vocabulary (OOV) rate is also provided for each domain.

| Domain/Dataset | Compared to SUBAK.KO train+val sets | | CNN | | wav2vec 2.0 XLS-R | | Whisper (multilingual) | |
|---|---|---|---|---|---|---|---|---|
| | # of OOV words | OOV rate (%) | CER | WER | CER | WER | CER | WER |
| Television news | 240 | 13.44 | 25.57 | 63.04 | 11.93 | 37.97 | 13.15 | 52.21 |
| Parliament speech | 228 | 14.38 | 40.44 | 75.97 | 14.06 | 40.55 | 14.77 | 56.37 |
| Audio books | 406 | 17.36 | 17.93 | 53.76 | 8.74 | 32.11 | 15.65 | 51.45 |
| Drama series | 288 | 17.69 | 41.72 | 79.79 | 19.07 | 48.56 | 17.31 | 55.00 |
| Class lecture | 213 | 16.54 | 28.82 | 65.52 | 12.46 | 37.72 | 17.64 | 51.65 |
| Political talk show | 238 | 13.24 | 27.99 | 64.82 | 13.50 | 40.41 | 12.41 | 57.82 |
| Celebrity Interview | 275 | 15.70 | 25.15 | 60.09 | 11.54 | 34.66 | 12.92 | 49.06 |
| Documentary | 262 | 14.78 | 24.93 | 61.20 | 10.19 | 32.48 | 14.05 | 51.93 |
| Kids' voice | 247 | 16.50 | 59.62 | 86.11 | 19.63 | 50.75 | 21.56 | 54.77 |
| Medicine | 446 | 34.47 | 34.85 | 74.96 | 14.51 | 42.40 | 18.72 | 63.86 |
| Biography | 318 | 18.29 | 12.79 | 43.07 | 7.78 | 28.25 | 10.76 | 56.18 |
| Kids' Cartoon | 247 | 17.16 | 48.55 | 81.46 | 23.65 | 54.17 | 28.26 | 66.86 |
| Sports | 377 | 21.13 | 32.06 | 69.40 | 17.30 | 47.37 | 13.30 | 61.67 |
| BanSpeech | 3585 | 30.83 | 31.84 | 67.65 | **14.02** | **40.62** | 15.94 | 55.92 |
| SUBAK.KO test | 918 | 5.24 | 4.02 | 13.94 | **2.02** | **7.56** | 4.96 | 22.15 |

All the models struggle recognizing speeches from domains like kids' voice, cartoon, medicine, drama, and sports. These domains containing spontaneous speeches are not included in the SUBAK.KO as this dataset is mostly a read speech corpus.

5.2 Experiments with CNN

Although a fully supervised CNN performs poorly on out-of-domain and spontaneous data, training a CNN is still a parameter-efficient way compared to fine-tuning large-scale pre-trained models such as wav2vec 2.0 and Whisper. CNN is applicable especially for low-resource and mid-resource languages with limited availability of graphics processing units. We conduct several experiments with CNN which is described in this subsection.

Three input feature extraction methods from audio signals such as MFCC, MFSC and Power Spectrum are implemented to build Bangla ASR models trained with our baseline deep CNN network and CERs/WERs are compared. Figure 1 illustrates the outcome of this experiment. MFCC technique gets slightly lower WER of 22.34% compared to MFSC which obtains 22.39% WER. Comparing the CERs, however, MFSC gets a slight edge with 6.34% CER than MFCC with 6.38% CER. On the other hand, using Power Spectrum features, the ASR model performs worse, getting a WER of 23.21% and a CER of 6.73%.

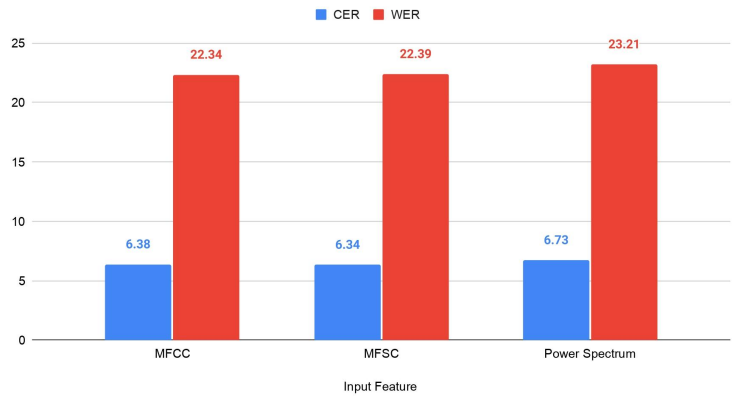

Fig. 1. ASR performance evaluation in terms of CERs and WERs calculated using the SUBAK.KO test set for the three input feature extraction methodologies such as MFCC, MFSC and Power Spectrum

Since we get lower WER using MFCC, we conduct further experiments with it by extracting different numbers of MFCCs from each frame. The WERs are calculated using SUBAK.KO dev and test sets and shown in Figure 2. We observe substantial variation in WERs while altering the number of MFCCs from 10 to 24 per frame. Initially, with increasing the MFCCs from 10 to 21, we can constantly improve the WERs from 26.20% to 22.34%. Taking 24 MFCCs per frame, however, slightly degrades the ASR performance with 22.53% WER. Similar phenomenon is observed on the dev set as well. These results suggest that a higher number of MFCCs (e.g. 21 MFCCs) generally ensures lower WER in the case of clean read speech training data.

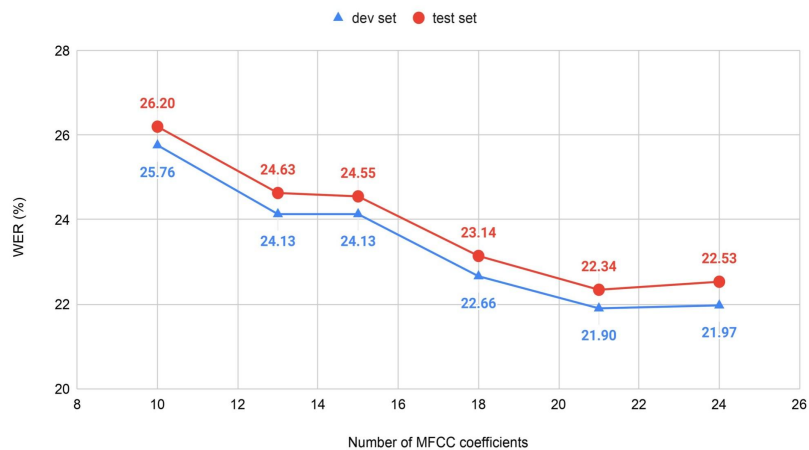

Fig. 2. Impact of number of MFCC coefficients on WER. SUBAK.KO dev and test sets are used for evaluation.

Table 3 presents the impact of windowing on CERs and WERs for a deep CNN-based ASR model. As seen from Table 3, frame size and frame stride have a strong impact on the ASR performance as well as the computational cost. Setting the frame size to 30 milliseconds (ms) and stride to 20 ms provides the lowest WER of 20.51% and CER of 6.04%. This setting also reduces the training time to only 17.44 hours. Based on this experiment, we suggest experimenting with the frame sizes and strides to find the suitable setting for each dataset while developing an ASR system.

**Table 3.** Impact of windowing on CERs and WERs calculated using an acoustic model trained with deep CNN and evaluated on the SUBAK.KO test set

| Frame size (in ms) | Frame stride (in ms) | CER | WER | Training Time (in hours) |
|---|---|---|---|---|
| 25 | 10 | 6.63 | 23.15 | 31.11 |
| 25 | 20 | 6.21 | 20.76 | 17.44 |
| 30 | 10 | 6.62 | 23.12 | 31.11 |
| 30 | 15 | 6.06 | 20.87 | 22.56 |
| 30 | 20 | **6.04** | **20.51** | **17.44** |
| 30 | 25 | 6.89 | 22.99 | 15.22 |

**Table 4.** The correlation of training dataset size and the number of convolutional layers measured on CER/WER using SUBAK.KO test set

| Hours of training data | CNN 10 Layers | | CNN 15 Layers | | CNN 20 Layers | |
|---|---|---|---|---|---|---|
| | CER | WER | CER | WER | CER | WER |
| 40 | 18.17 | 54.30 | 16.45 | **47.94** | 18.67 | 52.32 |
| 80 | 11.84 | 38.72 | 10.81 | 34.16 | 11.14 | 34.07 |
| 120 | 9.43 | 31.57 | 8.09 | 26.13 | 7.95 | 25.03 |
| 160 | 8.13 | 27.66 | 6.79 | 22.02 | 6.33 | 19.94 |
| 200 | 7.27 | 25.04 | 5.86 | 19.23 | 5.61 | **17.79** |

The performance of Neural Networks rest on the amount of data and the complexity of the networks/number of parameters. We split the SUBAK.KO train set and prepare 5 subsets containing 40 hours, 80 hours, 120 hours, 160 hours, and 200 hours of speech data. Using each of the subsets, we train three ASR models with 10, 15, and 20 convolutional blocks. The correlation between training data size and number of convolutional layers is shown in Table 4. In the case of 40 hours of training data, the CNN model with 15 convolutional layers outperforms the models with 10 layers and 20 layers, getting a WER of 47.94% and CER of 16.45%. The advantage of the deep CNN with 20 convolutional layers becomes more apparent as we increase the amount of data. Using 200 hours of train data, CNN with 10, 15, and 20 convolutional layers obtain 25.04%, 19.23%, and 17.79% WERs, respectively.

In Table 5, we provide the CERs and WERs on the SUBAK.KO test set for several types of normalization techniques integrated to our CNN model with 15 convolutional blocks. The normalization methods investigated in this study are batch norm [Ioffe and Szegedy 2015], layer norm [Ba et al. 2016] and weight norm [Salimans and Kingma 2016]. To implement that, the normalization method is applied to each convolutional layer, precisely after the dropout layer in each convolutional layer. Here, baseline CNN refers to the model where no normalization is applied. By adopting normalization, our model achieves lower WERs and CERs compared to our baseline. Using layer norm, we get 18.89% WER whereas applying batch norm and weight norm, 19.78% and 20.32% WERs can be obtained, respectively. While training a CNN with both batch norm and weight norm, our ASR model is able to get 20.14% WER. By building a model incorporating all of these normalization techniques (e.g. layer norm, batch norm, and weight norm), we achieve 19.29% WER. From this experiment, it can be concluded that CNN with layer norm outperforms the rest of the models.

**Table 5.** CER/WERs calculated on the SUBAK.KO test set for several normalization techniques. Here, B.N., W.N., and L.N. refers to batch normalization, weight normalization, and layer normalization, respectively.

| Model | Test CER | Test WER |
|---|---|---|
| Baseline CNN | 6.04 | 20.51 |
| CNN+batch norm. | 5.83 | 19.78 |
| CNN+weight norm. | 5.95 | 20.32 |
| CNN+layer norm. | 5.41 | **18.89** |
| CNN+B.N.+W.N. | 5.96 | 20.14 |
| CNN+L.N.+B.N.+W.N. | 5.55 | 19.29 |

# 6 LIMITATIONS

The speech utterances for BanSpeech have been collected from publicly available sources using an automated pipeline and so any sort of bias such as gender bias, political bias, etc. cannot be identified and filtered. Yet, we strictly monitor the domains from which we extract our data to build this benchmark. Another limitation is that gender parity cannot be determined in our dataset in terms of number of speakers. However, due to the random selection of audio sources within a domain from YouTube, we expect a significant number of utterances from both male and female speakers.

We collect speech data for the dialectal domain and use Google STT to transcribe them. In our initial evaluation, the quality of automated annotation of dialectal data is quite poor and we are not able to label them under human supervision as these are highly deviant from standard Bangla. These speeches, however, can be used for the language/dialect identification task.

All the domains contain approximately 30 minutes of annotated speech data. Due to resource constraints (e.g., time and cost), the size of these domains cannot be expanded. However, if resources were available, expanding them would be a promising next step.

wav2vec 2.0 XLS-R is pre-trained on a large amount of cross-lingual data. Although Whisper is also cross-lingual, it mostly contains English data and most of the other languages have less than 1000 hours of data [Radford et al. 2022]. If Whisper had more cross-lingual data, it would have been a better comparison between self-supervised and weakly supervised-based pre-training approaches. However, to the best of our knowledge, Whisper is the only large-scale weakly supervised pre-trained speech model as of now that can be compared to XLS-R. Moreover, both Whisper and XLS-R contain Bangla data in their pre-training dataset and so we make use of them in our work.

# 7 CONCLUSION AND FUTURE SCOPE

In this work, we introduce a multi-domain Bangla ASR evaluation benchmark, named BanSpeech, consisting of 6.52 hours of human-annotated speech data and 8085 utterances from 13 distinct domains. This dataset has been utilized to evaluate three state-of-the-art ASR training strategies including full supervision, self-supervision, and weak supervision. We use SUBAK.KO, which was developed by collecting recording scripts from 40 text domains following the reception and production criteria set for text domains for training/fine-tuning. Through a comprehensive evaluation, we find that a self-supervised, pre-trained, cross-lingual wav2vec 2.0 is considerably robust for recognizing out-of-domain speech data. Although Whisper has seen only 1.3 hours of Bengali data during weakly supervised pre-training, the performance does not improve even after fine-tuning it with 200 hours of high-quality annotated speech data. A CNN, trained from scratch, is found to be performing quite poorly in terms of robustness and generalizability. Thus, transfer learning or self-supervised speech models, to be more specific, is the best strategy to handle challenging spontaneous, noisy, and multi-domain speech recognition.

Empirical evaluation also suggests that SUBAK.KO-based CNN can perform comparatively better in domains containing mostly read speech, such as audiobooks, and biography. On the other hand, this model is not well-trained for perceiving spontaneous speech, as seen by its high WERs in domains such as drama, talk shows, sports, etc. Although 40 text domains were considered while building the mostly read speech corpus SUBAK.KO, these are not sufficient to address the complexities of spontaneous speech. We emphasize the requirement of compiling a spontaneous Bangla speech corpus considering distinct domains such as medicine, sports, children's voice, etc. for better ASR performance.

This paper also reports the experimental results on feature extraction techniques, normalization methods integrated into the deep CNN architecture, and the impact of convolutional layers in respect of different training data sizes. We find that the MFCC features with a high number of coefficients, the layer normalization strategy, and stacking many convolutional blocks can improve the performance of CNN-based ASR.

We would like to build a high-quality spontaneous Bangla speech corpus consisting of diverse domains as part of our future work. Since self-supervised models are proven to be much better for ASR under challenging conditions, we aim to conduct more research into them. More specially, there does not exist any mono-lingual Bangla wav2vec 2.0 and so we would like to pre-train a large-scale Bangla wav2vec 2.0 using both read speech and spontaneous speech utterances and compare it with the XLS-R for numerous speech processing tasks such as ASR, language identification, keyword spotting, etc. We are going to release BanSpeech and the CNN, wav2vec 2.0, and Whisper models for public use to foster research in Bangla.

## Appendix A

**Table 6** Hyper-parameters and other experimental choices used for training the ASR models

| Hyper-parameter | CNN | wav2vec 2.0 XLS-R | Whisper |
| --- | --- | --- | --- |
| Learning rate | 0.05 | 3e-5 | 1e-5 |
| Warm up steps | - | 300 | 500 |
| Max grad norm | 0.2 | - | - |
| Momentum | 0.8 | - | - |
| lrcrit | 0.006 | - | - |
| Cosine scheduler | False | True | True |
| Batchsize | 16 | 1 | 4 |
| Gradient accumulation | 1 | 2 | 1 |
| Max epochs | 200 | 30 | 15 |
| Early stopping patience | - | 10 | 10 |
| Seed | 100 | 42 | 42 |
| Token type | char | char | char |
| Pre-train model | - | XLS-R 300M | Medium 769M |